%% file: main.tex
\documentclass[11pt]{article}
\usepackage{microsoft-tech-report}
\usepackage{amsmath,amssymb,mathtools}
\usepackage{graphicx,booktabs,array,tabularx}
\usepackage{xcolor}
\usepackage{microtype}
\usepackage{placeins}
\usepackage{tcolorbox}
\usepackage{hyperref}
\usepackage{url}
\usepackage{needspace}
\titleformat*{\section}{\Large\msftsans\bfseries\color{msftdark}\raggedright}
\titleformat*{\subsection}{\large\msftsans\bfseries\color{msftdark}\raggedright}
\hypersetup{
  colorlinks=true,
  linkcolor=msftblue,citecolor=msftblue,urlcolor=msftblue,
  pdftitle={EnvOpt: Closed-Loop Environment Optimization for Agent Post-Training},
  pdfpagelayout=OneColumn,
  pdfpagemode=UseNone,
  pdfstartview=Fit
}

\newcommand{\method}{\textsc{EnvOpt}}

\newcommand{\ind}{\mathbf{1}}

\newcommand{\code}[1]{\texttt{#1}}

\title{EnvOpt: Closed-Loop Environment Optimization for Agent Post-Training}
\author{Zisu Huang$^{1,*,\dagger}$ \hspace{0.1em}
  Zicong Zhang$^{3,4,*,\dagger}$ \hspace{0.1em}
  Jingwen Xu$^{1,*,\dagger}$ \hspace{0.1em}
  Rui Wang$^{1,*,\dagger}$ \hspace{0.1em}
  Yifan Yang$^{2,3,\ddagger}$ \hspace{0.1em}
  Muzhao Tian$^{1,\dagger}$\\[0.2em]
  Xuemei Gao$^{2}$ \hspace{0.1em}
  Qi Dai$^{2}$ \hspace{0.1em}
  Bei Liu$^{2}$ \hspace{0.1em}
  Kai Qiu$^{2}$ \hspace{0.1em}
  Dongdong Chen$^{2}$ \hspace{0.1em}
  Xiaoqing Zheng$^{1,\ddagger}$ \hspace{0.1em}
  Chong~Luo$^{2}$}
\techreportshorttitle{EnvOpt}

\makeatletter
\renewcommand{\maketitle}{\thispagestyle{empty}
  \noindent
  \begin{minipage}[c]{0.5\linewidth}
    \raggedright\raisebox{-0.5\height}{\msftbrandmark}
  \end{minipage}\begin{minipage}[c]{0.49\linewidth}
    \raggedleft{\msftdatefont\small\color{msftgray}December 2026}
  \end{minipage}\par
  \vspace{0.35em}
  \noindent{\color{msftline}\rule{\linewidth}{0.8pt}\par}
  \vspace{1.0em}
  \begin{center}
    {\msfttitlefont\fontsize{18}{22}\selectfont\color{msftdark}\@title\par}
    \vspace{1.25em}
    {\fontsize{9.5}{12}\selectfont\rmfamily\color{msftdark}\@author\par}
    \vspace{0.22cm}
    {\footnotesize\rmfamily\color{msftgray}
      $^{1}$ Fudan University \quad
      $^{2}$ Microsoft \quad
      $^{3}$ Shanghai Jiao Tong University \quad
      $^{4}$ Harvard University\par}
  \end{center}
  \vspace{0.45em}
}
\makeatother

\begin{document}
\maketitle
\suppressfloats[t]
\begin{msfttitlebox}
\setlength{\parindent}{0cm}
\setlength{\parskip}{0.14cm}
\raggedright
\hyphenpenalty=10000
\exhyphenpenalty=10000
\begin{abstract}
Data-centric recursive self-improvement (RSI) seeks to improve a model by repeatedly turning its evolving behavior into new training data. For tool-using agents, sustaining this process is more difficult because the data are executable environments: each round must preserve a coherent and tight connection from observed behavior to training intent, executable realization, and realized learning outcomes. Our analysis finds that even a capable agent frequently loses this connection, showing an incomplete understanding of model behavior, unfaithful implementation of intended environments, and ineffective use of feedback across iterations. To address these challenges, we introduce \method{}, a framework that organizes environment evolution around traceable training targets. It combines a traceable behavioral map to expose recurring patterns, iterative environment construction to faithfully realize training targets as executable environments, and target-conditioned context refinement to facilitate accumulated feedback utilization. This turns environment generation from a sequence of isolated attempts into a closed-loop data improvement process that adapts as the target model changes. Extensive experiments on $\tau^2$-Bench, BFCL-v4, and AppWorld show that \method{} yields substantial improvements in the target model's performance, while the vanilla agent baseline shows no consistent improvement trend.
\end{abstract}
\vspace{0.14cm}
{\setlength{\parskip}{0.06cm}\small
{\msftmetalabel{Correspondence}
\href{mailto:yifyang29@outlook.com}{yifyang29@outlook.com},
\href{mailto:zhengxq@fudan.edu.cn}{zhengxq@fudan.edu.cn}\par}
{\msftmetalabel{Code}
\href{https://github.com/microsoft/EnvOpt}{https://github.com/microsoft/EnvOpt}\par}
{\msftmetalabel{Webpage}
\href{https://microsoft.github.io/EnvOpt/}{https://microsoft.github.io/EnvOpt/}\par}
}
\vspace{0.08cm}
{\footnotesize\rmfamily\itshape\color{msftgray}
$^*$ Equal contribution. \quad
$^{\dagger}$ Work done during an internship at MSRA. \quad
$^{\ddagger}$ Corresponding authors.\par
}
\end{msfttitlebox}

\input{sections/introduction}

\input{sections/related_work}
\input{sections/evidence_gaps}
\input{sections/method}
\input{sections/experiments}
\input{sections/conclusion}
\clearpage

\section*{AI use statement}
In this work, we used generative AI tools to assist with methodological formulation, experimental design, method implementation, qualitative analysis of model behavior, and interpretation of results. Our research system also uses language models to construct training environments and tasks, simulate users, and generate and evaluate trajectory-analysis reports, as described in the paper. Additionally, we used generative AI for literature analysis, manuscript drafting and editing, and figure preparation. The authors take responsibility for the final content of this work, including text, claims, code, and figures produced with the assistance of generative AI.

\section*{Reproducibility statement}
We describe the data-research loop and method in Sections~\ref{sec:evidencegaps} and~\ref{sec:method}. Appendix~\ref{app:experimentalsetup} provides implementation details, training and evaluation settings, benchmark splits, and baseline implementations. Appendix~\ref{app:analysisexperiments} details the procedures and metrics used in our analysis experiments. All benchmark datasets used in our experiments are publicly available. We will release our code, including prompts, configurations, and data splits, to support reproducibility.

\section*{Ethics statement}
This work studies agent post-training using publicly available benchmarks and programmatically constructed environments. It does not involve human-subject experiments. Real-world deployment of tool-using agents should include appropriate access controls and privacy safeguards.

\bibliography{references}
\bibliographystyle{unsrtnat}

\clearpage
\appendix
\input{sections/appendix_limitations}

\input{sections/appendix_codex_vanilla_case_studies}

\input{sections/appendix_analysis_experiments}

\input{sections/appendix_experimental_setup}

\end{document}

%% file: sections/introduction.tex
\section{Introduction}
\label{sec:intro}

\input{figures/overview_float}

Recursive self-improvement (RSI) asks whether AI systems can use evidence from one development cycle to improve the next. MetaRSI~\citep{metarsi} identifies data, harness, and model as three writable surfaces for improvement. Among these, autonomously developing training data and using it to update model parameters offers a key pathway toward stronger intelligence. We therefore focus on the interaction between the data and model surfaces, where a data agent constructs and adapts training data as the target model evolves. RSIBench-Data~\citep{rsibenchdata} shows that agents can discover useful data strategies but struggle to improve them consistently across training runs, exposing a discovery--reliability gap. This raises a sharper question: when the target model changes, can feedback from one training round reliably guide what data to construct next?

We study this question in the context of training agents to use tools more effectively, as tool use is a fundamental capability for modern agents. Training such agents extends beyond generating static samples because training data take the form of executable instances that define state, tool behavior, tasks, and outcome verification. Systems for executable environment synthesis and scaling, including AgentScaler, EnvScaler, ScaleEnv, Agent World Model, and EnvFactory~\citep{agentscaler,envscaler,scaleenv,awm,envfactory}, show that such resources can be constructed automatically. These works establish the feasibility of automated environment construction, while data-centric RSI further requires the agent to decide how these environments should evolve as the target model changes. We instantiate this process in a stateful, multi-round online-RL setting, where a target model is iteratively updated using new data produced by a data agent. Starting from an empty environment pool, the agent repeatedly analyzes the current target model's behavior, constructs executable training instances, submits them to a fixed external service for online-RL training and evaluation, and uses the resulting feedback to guide the next round. With each round continuing from the updated model, the training data and target policy evolve together over time. Reliable recursive improvement is therefore not a one-shot data-generation problem: it requires sustaining a coherent data loop as both continue to evolve. Our experiments show that even a modern agent with strong reasoning and coding capabilities (Codex with GPT-6 Astra~\citep{openai2026gpt6astra}) struggles to sustain improvement across rounds.

Based on our experiments and observations, we identify three key stages in closing the data improvement loop and a recurring failure mode at each. Model behavior understanding is limited by \textbf{incomplete analysis} of available traces, yielding a partial or biased view of the target model. Executable data construction exhibits \textbf{unfaithful realization} when data instances do not exercise the intended training target. Cross-round feedback utilization is disrupted by \textbf{disconnected feedback}, which prevents the agent from relating constructed data to training dynamics and subsequent target behavior.

To address these challenges, we introduce \method{}, a framework that organizes environment evolution around traceable training targets through three corresponding components. To identify patterns in target model behavior, a \emph{traceable behavioral map} consolidates evidence across trajectories while preserving links to the source cases. To faithfully realize each training target as executable data, \emph{iterative environment construction} uses a planning, implementation, and refinement loop to bridge the last mile from target specification to logically consistent and behaviorally aligned environments. To close the feedback loop across rounds, \emph{target-conditioned context refinement} organizes relevant data, observed training dynamics, and subsequent evaluation outcomes around each training target.
Experiments on three tool-use benchmarks demonstrate that \method{} sustains multi-round improvement. Averaged across the three benchmarks, it improves performance by $+16.22$, $+7.40$, and $+6.36$ percentage points on Qwen3-4B, Qwen3-8B, and Qwen3-30B-A3B. Its performance shows a clear overall upward trend across iterations, while vanilla agents fluctuate without sustained improvement. Further analyses show improved task and failure coverage, more reliable and target-aligned environments, and higher-quality realized RL signals.

%% file: figures/overview_float.tex
\begin{figure}[t]
\centering
\includegraphics[width=\linewidth]{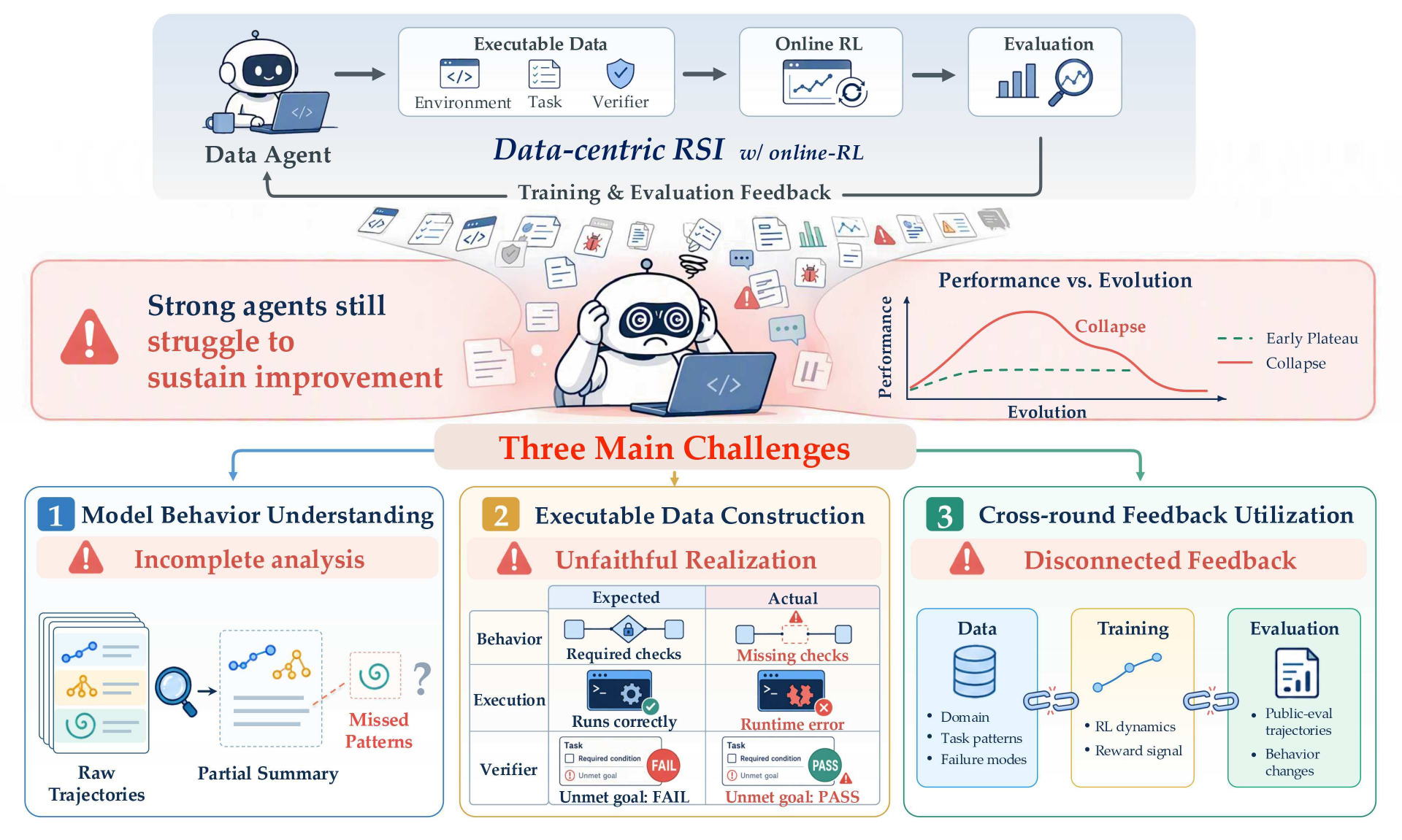}
\caption{Three challenges in closing the data-centric RSI loop: incomplete behavioral analysis, unfaithful realization of executable training instances, and disconnected data--training--evaluation feedback.}
\label{fig:overview}
\end{figure}

%% file: sections/related_work.tex
\section{Related Work}
\label{sec:related}

\paragraph{Data-centric RSI.}
RSIBench-Data studies agents that revise training-data strategies for a
separate target model under fixed training, evaluation, and resource
controls~\citep{rsibenchdata}. Autodata develops an agentic data scientist
that iterates between data creation and quality assessment, including
meta-optimization of the data scientist itself~\citep{autodata}. DataMaster
combines data discovery and transformation with tree search and cumulative
memory while holding the learning algorithm fixed~\citep{datamaster}.
\method{} studies this research role for executable training environments:
a researcher starts with an empty environment library and uses
behavioral evidence to guide successive rounds of environment revision and
online RL. Its focus is the evidence supporting those research decisions.

\paragraph{Executable environment synthesis for tool-calling agents.}
AgentScaler, EnvScaler, and ScaleEnv automate the construction of executable
tools, state, and verifiable tasks~\citep{agentscaler,envscaler,scaleenv}.
Agent World Model scales database-backed synthetic environments for RL,
while EnvFactory combines executable environment construction with
topology-aware task synthesis~\citep{awm,envfactory}. FACET grounds task
instructions, solutions, and verifiers in shared executable state to preserve
source intent and cross-artifact consistency~\citep{facet}. The ACE framework
and WRIT further examine data accuracy, learner-relative complexity,
diversity, and the evidence required for individual tool-use
decisions~\citep{ace,writ}. These works establish construction and validation
principles that \method{} builds on; our emphasis is using source-traceable
behavioral analysis and actual-training feedback to decide which mechanisms
to construct or revise.

\paragraph{Iterative RL and environment adaptation.}
PAIRED and Prioritized Level Replay adapt environment generation or selection
using regret and learning-potential signals~\citep{paired,plr}. For tool-use
agents, RODS uses online reward statistics to synthesize structurally aligned
tasks and update a replay pool~\citep{rods}. EnvHarness adapts existing
environments from policy trajectories while retaining their original
verifiers~\citep{envharness}. SPADE trains a shared environment-designer and
solver model through self-play, using hint-based regret, corpus grounding,
and environment memory~\citep{spade}. Agent-World and AgentOmnia also use
behavioral diagnosis to guide targeted task synthesis and
self-evolution~\citep{agentworld,agentomnia}. \method{} uses an external
researcher to revise independent environments without the target backend or
verifier. It connects the source evidence behind a diagnosis to realized
program versions, training tasks, RL observations, and fixed-target
evaluation, supporting inspection of what each intervention actually delivered.

%% file: sections/evidence_gaps.tex
\section{Agents as Data Researchers: Identifying Challenges in Data-Centric RSI}
\label{sec:evidencegaps}

We instantiate data-centric RSI as an iterative loop between a data agent and a target model. The agent constructs executable environments and tasks for training and uses feedback to guide subsequent data decisions. We first formalize this loop and then examine its key challenges.

\paragraph{Formalizing the data loop.}
We model an executable environment as a partially observable interaction process
\[
e=(\mathcal{S},\mathcal{A},\mathcal{T},\mathcal{O}).
\]
Here, $\mathcal{S}$ denotes the state space over structured environment data (e.g., database records), and $\mathcal{A}$ contains the executable actions exposed through tool calls. The transition function $\mathcal{T}(s' \mid s,a)$ determines how an action updates the state, while the observation function $\mathcal{O}(o \mid s',a)$ returns the tool output or user response visible to the target model.

At round $t$, the data agent constructs a set $D_t=\{(e_j,x_j,v_j)\}_{j=1}^{n_t}$ of $n_t$ executable training instances. The task $x_j$ defines the goal and initial conditions, while the verifier $v_j$ evaluates task completion. The data generation process is conditioned on a research context $C_t$, which typically includes the agent's execution history, accumulated memory, prior feedback, external examples, and the current pool of environments and tasks. Formally,
\begin{equation}
\label{eq:researchoutput}
D_t=\operatorname{DataAgent}(C_t).
\end{equation}

Let $\pi_t$ denote the target model at the start of round $t$. $D_t$ is submitted to a fixed external service, which updates $\pi_t$ through online RL and evaluates the resulting model:
\begin{equation}
\label{eq:researchloop}
(\pi_{t+1},F_t)=\operatorname{TrainEval}(\pi_t,D_t),
\end{equation}
where $\pi_{t+1}$ is the updated target model and $F_t$ denotes the resulting feedback, comprising training dynamics and evaluation results. Finally, the generated data and resulting feedback are incorporated into the research context for the next round:
\begin{equation}
\label{eq:contextupdate}
C_{t+1}=\operatorname{UpdateContext}(C_t,D_t,F_t).
\end{equation}

The loop starts from an empty environment pool, and each subsequent round continues from the updated model $\pi_{t+1}$, allowing the generated data and target model to evolve together. To better understand this process, we analyze three critical aspects of the loop and identify common failure modes in each, focusing on how the data agent uses $C_t$ to understand model behavior, realizes the selected training targets as executable instances in $D_t$, and incorporates $F_t$ into $C_{t+1}$.

\subsection{Model Behavior Understanding: Incomplete Analysis}
\label{sec:coveragegap}

Using $C_t$ to select suitable training targets requires a comprehensive understanding of the target domain, task patterns, model behavior, and recurring failures. To evaluate this capability, we measure the behavioral coverage of Codex with GPT-6 Astra when analyzing raw trajectories. We give the agent full access to the trajectories and ask it to produce a comprehensive report on the observed task patterns and failure modes. To promote a fair comparison and reduce evaluator self-preference, we use Gemini 3.8 Flash~\citep{google2026gemini38flash}, from a different model family, to construct evidence-grounded reference units and evaluate the report against them. Appendix~\ref{app:behavioralcoverage} provides the full evaluation details. Across $1{,}860$ trajectories from three benchmarks, the report achieves average task-pattern and failure-mechanism coverage of $78.1\%$ and $58.8\%$, respectively. These results show that the agent may overlook important cases and therefore miss the behavioral evidence needed to determine what training data the target model requires (Figures~\ref{fig:codex_read_write} and~\ref{fig:coverage-exmaples}).

\subsection{Executable Data Construction: Unfaithful Realization}
\label{sec:implementationgap}

Executable data construction requires environment behavior, task requirements, and verifier feedback to jointly and faithfully realize the intended training target. Yet executability alone does not guarantee faithful realization. The cases in Appendices~\ref{app:codexunfaithful} and~\ref{app:implementationgapcases} illustrate three failure modes concerning what behavior is trained, how the interaction operates, and how task completion is judged. \emph{Missing task mechanisms} allow the model to bypass the intended behavior. \emph{Faulty interaction logic} causes responses, state transitions, or termination conditions to violate the task rules. \emph{Misaligned verifiers} produce rewards that do not accurately reflect the model's progress or satisfaction of the task requirements. These failures produce training experience that deviates from the intended target, making subsequent feedback unreliable for guiding environment evolution.

\subsection{Cross-round Feedback Utilization: Disconnected Feedback Across Data, Training, and Evaluation}
\label{sec:feedbackgap}

Producing suitable, learnable, and capability-aligned tasks requires tight connections among targeted data creation, actual training, and subsequent evaluation outcomes. By integrating information from these three stages, the agent can identify which capabilities require improvement and adjust its data generation strategy accordingly. However, even strong agents struggle to identify the most suitable directions and strategies for data generation (Figure~\ref{fig:codex_reward_saturation} shows a real case where task difficulty remains unchanged despite saturated rewards). We observe that only $21.4\%$ of the training tasks generated by the vanilla Codex agent exhibited reward variation across rollouts (Figure~\ref{fig:researchdiagnostics}). This limited proportion suggests that feedback was not effectively translated into learnable training tasks, leaving much of the available training signal underutilized.

%% file: sections/method.tex
\section{Method}
\label{sec:method}

\method{} instantiates the data-research loop in
Equations~\ref{eq:researchoutput}--\ref{eq:contextupdate}
(Figure~\ref{fig:framework}) through $3$ components corresponding to the
challenges in Section~\ref{sec:evidencegaps}. A \emph{traceable behavioral map}
(TBM) preserves task requirements and failure mechanisms with links to source
trajectories. Guided by the research context $C_t$,
\emph{iterative environment construction} (IEC) produces the
environment--task--verifier set $D_t$ through planning, implementation, and
independent review. \emph{Target-conditioned context refinement} (TCR)
organizes training targets, actual RL experience, and evaluation feedback
to refine this context for the next iteration. The fixed training-and-evaluation
service updates the target model and returns feedback for the next iteration.

\input{figures/framework_float}

\subsection{Traceable Behavioral Map}
\label{sec:diagnosis}

The map separates a persistent evidence layer from its compact reading view.
An LLM performs extraction, source review, merging, and refinement.
Let $T_{t,i}$ contain the raw trajectories for task $i$, including
instructions, tool definitions, and observed interactions. Extraction and
source review produce a set of source-grounded mechanism records:
\begin{equation}
\mathcal M_t=\bigcup_i \operatorname{ExtractReview}(T_{t,i}).
\label{eq:mechanismledger}
\end{equation}
Each record links an environment mechanism, task requirement, or observed
failure or recovery to its source and applicable conditions.
Bounded hierarchical merging builds a draft by factoring shared premises
while retaining distinct actors, scopes, dependencies, and recovery branches:
\begin{equation}
R_t^{\mathrm{draft}}=\operatorname{FactorMerge}(\mathcal M_t).
\label{eq:factoredmap}
\end{equation}
For refinement, we randomly sample $10\%$ of the mechanism records to form
a review subset $\mathcal M_t^{\mathrm{check}}\subseteq\mathcal M_t$.
Checking only this compact subset makes refinement cost-efficient by avoiding
reanalysis of complete trajectories.
An LLM checks whether the draft preserves the sampled mechanisms and their
applicable conditions. $\operatorname{Repair}$ corrects conflicting report
entries, and $\operatorname{Supplement}$ consolidates missing distinctions
identified in the sampled records:
\begin{equation}
\begin{aligned}
\bar R_t &= \operatorname{Repair}(R_t^{\mathrm{draft}},\mathcal M_t^{\mathrm{check}}),\\
R_t &= \bar R_t \oplus \operatorname{Supplement}(\mathcal M_t^{\mathrm{check}},\bar R_t).
\end{aligned}
\label{eq:residualmap}
\end{equation}
Here $\oplus$ appends the missing distinctions without rewriting preserved
content. An already-covered record must instead be matched to an existing
passage.

The report retains the four fields \emph{domain, environment, task behavior,
failure behavior}. It enters the research context $C_t$ with links back to
original trajectories. Section~\ref{sec:analysiscoverage} evaluates the
behavioral coverage of the resulting reports.
Appendix~\ref{app:componentimplementation} provides an example of the generated report.

\subsection{Iterative Environment Construction}
\label{sec:environment}

Using the research context $C_t$, which integrates behavioral analysis,
training feedback, and accumulated experience, the data agent selects
environments to create or revise and allocates training tasks. It specifies
each environment's objective, requirements, and expected behaviors, then
delegates implementation to a persistent Env Builder subagent.
\label{sec:checks}
Each builder produces candidate $(e,x,v)$ triples aligned with the specification.
The current target model then performs probe rollouts on representative tasks.
An independent review agent examines the implementation, generated tasks,
execution checks, and probe rollouts to assess implementation correctness,
target alignment, and potential task difficulty. It produces a review report
with a decision to accept the candidate or request refinement, together with
actionable feedback. Implementation issues return to the same builder for revision
and re-review, while specification or difficulty issues return to the data agent
for subsequent data decisions.
Finally, this process produces a batch of quality-checked training data $D_t$
for RL. By aligning environment behavior and reward feedback with the intended
capability targets, iterative refinement provides more reliable learning
signals for the target model.

\subsection{Target-Conditioned Context Refinement}
\label{sec:planning}

Effective data evolution requires interpreting training and evaluation feedback
in relation to the capabilities the generated data was intended to improve.
We therefore organize the research context around explicit capability targets.
Before proposing environment changes, the data agent records the target
requirement and expected effect based on observed model behavior. These
provide a shared reference for interpreting subsequent feedback and deciding
what data to generate next.

We then perform version-aware evidence alignment. Environment versions and
training-task IDs link each target to its accepted instances in $D_t$ and
actual RL groups and rollouts. Associated evaluation tasks connect these
training records to performance on the targeted capability. The resulting
target-conditioned evidence view distinguishes planned, accepted, and actually
trained data while bringing together three complementary perspectives:
\begin{equation}
\underbrace{\text{Training target}}_{\text{expected capability}}
\longleftrightarrow
\underbrace{\text{Actual training}}_{\text{RL dynamics}}
\longleftrightarrow
\underbrace{\text{Target evaluation}}_{\text{related performance}}.
\label{eq:lineage}
\end{equation}

Context refinement proceeds through expectation--observation comparison.
The data agent jointly examines training rewards, within-task reward contrast,
and related evaluation outcomes and trajectories against the expected effect.
It assesses whether the generated data provided useful learning signals and
whether the targeted behavior improved, recording supporting observations,
mismatches, and unresolved questions in $C_{t+1}$. These accumulated findings
guide revisions to environment mechanisms, task difficulty, and sampling,
as well as decisions to retain or pause environments. Each iteration thus
refines the research context used to plan the next training intervention.

%% file: figures/framework_float.tex
\begin{figure}[t]
\centering
\includegraphics[width=\linewidth]{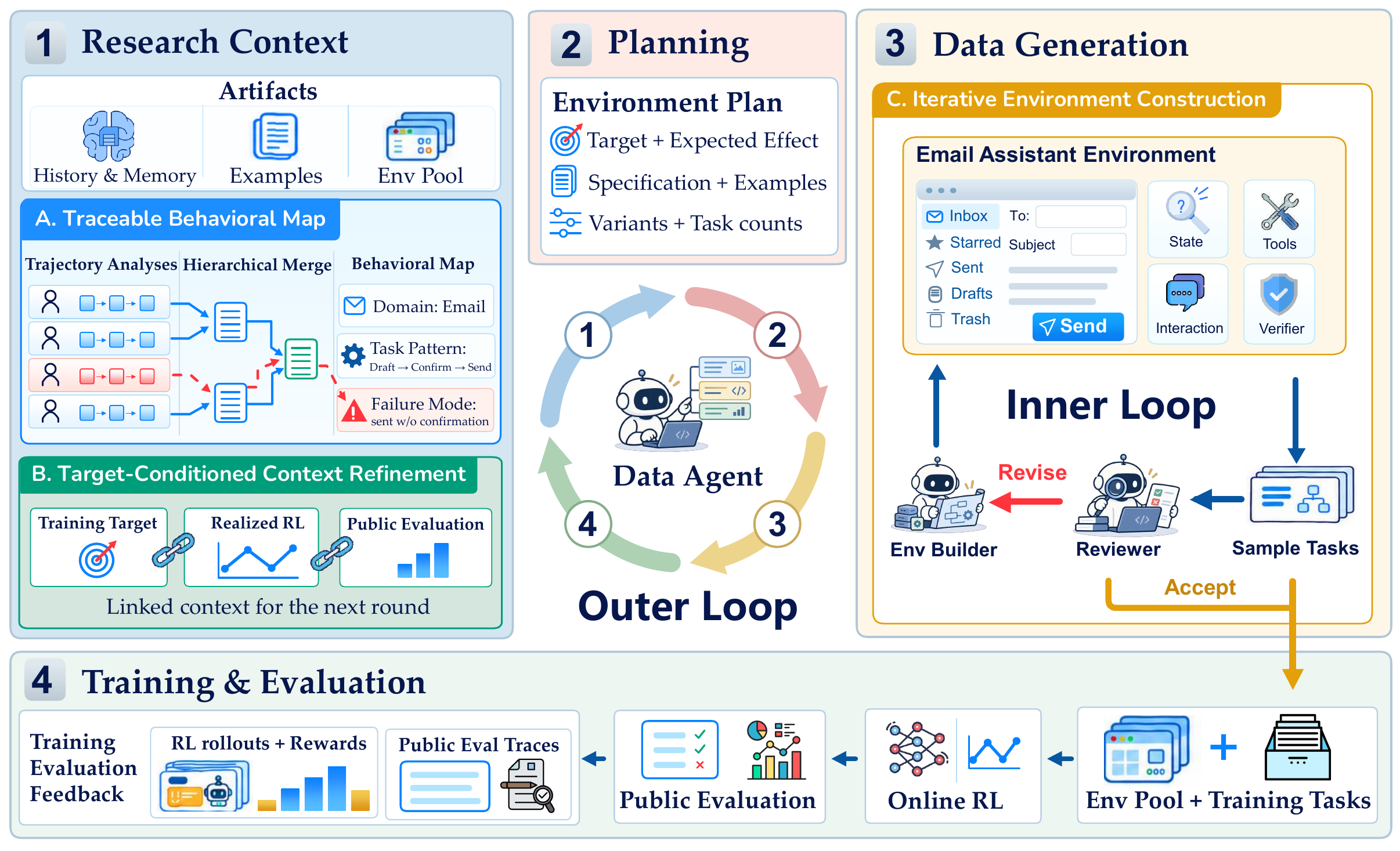}
\caption{The evolution loop of \method{}. The data agent analyzes model behavior, plans and constructs training environments, and uses feedback from online RL and public evaluation to guide the next iteration.}
\label{fig:framework}
\end{figure}

%% file: sections/experiments.tex
\section{Experiments}
\label{sec:experiments}

\subsection{Experimental Setup}

\paragraph{Evaluation and training.}
We evaluate all methods on $\tau^2$-Bench~\citep{tau2}, BFCL-v4 (multiturn)~\citep{bfcl}, and AppWorld~\citep{appworld}. $\tau^2$-Bench and BFCL-v4 use $1{:}1$ public and private splits, while AppWorld uses its train split for public evaluation and its test-normal split for private evaluation. Results on the public splits guide environment evolution, whereas the private splits remain hidden throughout the evolution process and are used only for final performance comparisons. We train the target model using GRPO~\citep{grpo} and report the performance of the final checkpoint produced by the complete evolution process on the held-out private evaluation splits. Reported benchmark scores are averaged over $3$ evaluation repeats.

\paragraph{Implementation and baselines.} Our target models are Qwen3-4B-Instruct-2507, Qwen3-8B, and Qwen3-30B-A3B-Instruct~\citep{qwen3}, with reasoning enabled only for Qwen3-8B. We instantiate the data agent, $\operatorname{DataAgent}$, with Codex and provide it with access to the GPT-5.5~\citep{openai2026gpt55} API and four A100 GPUs during evolution. We compare against vanilla Codex acting as an autonomous data agent (\texttt{Codex-Vanilla}) and online RL using a fixed environment set generated once by Codex without subsequent revision (\texttt{Fixed-Env}). We also compare with two static environment-scaling methods, \texttt{AWM}~\citep{awm} and \texttt{EnvScaler}~\citep{envscaler}. For each method, we train the target model on 3K tasks sampled from its publicly released environments and tasks. This is about three times the 960-task budget of \method{}, and the data remain fixed throughout training. All data-agent configurations use Codex powered by GPT-6 Astra with high reasoning effort. Further experimental details are provided in Appendix~\ref{app:experimentalsetup}.

\subsection{Main Results}
\label{sec:mainresults}

We report performance on the held-out private evaluation splits for
\method{} and all baselines in Table~\ref{tab:mainresults}.
Figure~\ref{fig:evolution} shows the
public-evaluation curves of \method{} and \texttt{Codex-Vanilla} across
evolution iterations.

\input{tables/benchmark_results}

\input{figures/evolution_float}

\paragraph{Vanilla strong agents struggle with continual data evolution.}
Despite pairing GPT-6 Astra with the Codex harness, \texttt{Codex-Vanilla}
does not deliver consistent improvements across benchmarks. Its average test
performance falls below the untrained model by $1.74$~pp for 4B, $7.03$~pp
for 8B, and $1.16$~pp for 30B, even though some individual benchmarks improve. The public-evaluation curves likewise show
early gains giving way to plateaus or regressions, including sharp late-stage
declines on $\tau^2$-Bench and AppWorld with 8B and on AppWorld with 30B.
Consistent with the challenges in Section~\ref{sec:evidencegaps},
these outcomes show that strong agent capabilities alone do not ensure that generated environments
remain useful as the target model changes.

\paragraph{\method{} yields substantial gains across three benchmarks.}
\method{} achieves the highest average test performance at all three model sizes.
With 4B, it reaches $38.30\%$ and exceeds \texttt{Codex-Vanilla} by $17.96$~pp.
The 8B and 30B models reach $36.03\%$ and $38.54\%$, exceeding the same baseline by $14.43$~pp and $7.52$~pp.
It improves over the untrained model on all three benchmarks at every size,
showing that the benefits extend across different tool-use domains.
The evolution curves also show that \method{} generally maintains and
extends gains across iterations, with especially
clear upward trends on 4B $\tau^2$-Bench and AppWorld.

\paragraph{Adaptive data evolution outperforms fixed environments.}
\texttt{Fixed-Env} improves average test performance over the untrained model
at 8B and 30B and roughly matches it at 4B, showing that a fixed set of generated environments can provide useful training data. However, \method{} achieves higher average performance at every size, with margins of $16.20$~pp for 4B, $3.83$~pp for 8B, and $3.85$~pp for 30B; it trails \texttt{Fixed-Env} only on 30B AppWorld. The static environment-scaling baselines show a similar pattern. Even with about three times the task budget of \method{}, \texttt{AWM} and \texttt{EnvScaler} bring little or no average gain at any model size. Generating a large set of environments once is therefore not enough; the training data need to keep adapting to the target model.

\subsection{Analysis}
\label{sec:analysiscoverage}

\input{figures/analysis_coverage_float}

\paragraph{Improved coverage of task patterns and failure mechanisms.}

We isolate the agent's ability to understand task patterns and recurring failures, directly examining the analysis gaps identified in Section~\ref{sec:coveragegap}. Following the same experimental setting, we compare the analysis produced by \texttt{Codex-Vanilla}, configured identically to the agent evaluated there, with our traceable behavioral map using the same trajectories and task and failure coverage metrics. As shown in Figure~\ref{fig:researchdiagnostics}a, our method consistently improves both metrics across all benchmarks, with average gains of $18.4$~pp in task coverage and $23.2$~pp in failure coverage. The largest improvements occur on BFCL-v4, reaching $26.6$~pp and $31.9$~pp, respectively. Our traceable behavioral map is also between $12\%$ and $29\%$ shorter than the vanilla agent's analysis. Figure~\ref{fig:coverage-exmaples} shows examples where the vanilla analysis omits a recovery mechanism or mischaracterizes a failure. Despite using the weaker GPT-5.5, our analysis is more complete and compact.

\paragraph{Sustaining learning signals across iterations.}
\label{sec:analysissignal}

Section~\ref{sec:feedbackgap} shows that the vanilla agent often fails to translate training and evaluation feedback into learnable task designs. Continual improvement requires tasks in the target model's learning region, where GRPO rollouts yield varying rewards. Figure~\ref{fig:researchdiagnostics}b reports the proportion of tasks that retain such reward variation in each round. Across all three benchmarks and evolution iterations, \texttt{Codex-Vanilla} generates tasks with reward variation at an average rate of only $21.4\%$. This rate even approaches $0\%$ in later rounds of $\tau^2$ and BFCL-v4. In contrast, \method{} achieves an average learnable task rate of $58.5\%$ under the same setting, yielding an absolute increase of $37.1$~pp and a $2.7\times$ improvement over \texttt{Codex-Vanilla}. Linking tasks to their actual training outcomes helps the agent detect reward saturation and redirect generation toward capabilities that remain learnable.

\paragraph{Environment implementation quality.}
\label{sec:analysisenvironment}
Our case studies illustrate the 3 types of construction problems defined in Section~\ref{sec:implementationgap}: missing task mechanisms, faulty interaction logic, and misaligned verifiers. In these cases, \method{}'s review-and-refinement loop resolves the identified defects by restoring required clarification, correcting interaction termination, and aligning rewards with complete, authorized outcomes. The resulting environments more faithfully implement their training targets, as shown in Appendix~\ref{app:implementationgapcases} (Figures~\ref{fig:env_refinement_interaction}--\ref{fig:env_refinement_verifier}).

\subsection{Ablation Study}
\label{sec:ablation}

Using Qwen3-4B-Instruct-2507, we enable each of the 3 proposed components
in isolation, yielding \texttt{TBM-only} (traceable behavioral map),
\texttt{TCR-only} (target-conditioned context refinement), and
\texttt{IEC-only} (iterative environment construction). We compare their test performance with
\texttt{Codex-Vanilla} and full \method{} across all 3 benchmarks.
Figure~\ref{fig:ablations} shows that all 3 single-component variants
outperform \texttt{Codex-Vanilla}, indicating that each component is useful
on its own. Their relative strengths vary: \texttt{TCR-only} performs best
among the variants on $\tau^2$-Bench, and \texttt{TBM-only} on BFCL-v4 and
AppWorld. Full \method{} nevertheless
achieves the best performance on every benchmark, exceeding the strongest
single-component variant by $1.98$--$6.00$ pp. These results
suggest complementary benefits from improving behavioral
understanding, feedback utilization, and environment construction.
Ablation details can be found in Appendix~\ref{app:ablationdetails}.

\input{figures/ablation_float}

\Needspace{9cm}
\subsection{Joint Optimization Across Benchmarks}
\label{sec:jointoptimization}

We further test whether \method{} can evolve one shared training set that jointly optimizes $\tau^2$-Bench, BFCL-v4, and AppWorld, using their public splits as feedback. ACEBench~\citep{acebench} and API-Bank~\citep{apibank} serve only as out-of-distribution (OOD) benchmarks (Figure~\ref{fig:jointoptimization}).

\input{figures/joint_optimization_float}

\paragraph{\method{} improves all optimized benchmarks together.}
Although the data agent must now balance different tool-use needs within one training set, \method{} improves test performance on all three optimized benchmarks at both model sizes. The gains range from $4.80$ to $11.31$~pp for 4B and from $4.36$ to $8.67$~pp for 8B, giving average gains of $7.56$~pp and $6.58$~pp, respectively. \texttt{Codex-Vanilla} is less stable: it drops below the untrained model on two benchmarks with 4B, and its 8B average gain of $3.17$~pp is about half that of \method{}. The dev curves also show that \method{} usually holds its gains, while \texttt{Codex-Vanilla} often stalls or slips back.

\paragraph{The gains carry over to unseen benchmarks.}
The OOD results suggest that the data agent does not simply fit the feedback benchmarks: \method{} improves on both held-out benchmarks at both model sizes. On ACEBench, the gains are $0.56$~pp for 4B and $4.17$~pp for 8B. On API-Bank, they are $5.87$~pp and $1.13$~pp. These benchmarks provide no feedback, so the learned tool-use behavior transfers.
\FloatBarrier

%% file: tables/benchmark_results.tex
\begin{table}[t]
\centering
\caption{Test performance (\%). \texttt{AWM} (3K) and \texttt{EnvScaler} (3K) are trained on 3K tasks sampled from each method's released data. Deltas from None (pp) are \textcolor[HTML]{218C4E}{green} for gains and \textcolor[HTML]{C43C39}{red} for drops. Avg. averages the three displayed benchmarks.}
\label{tab:mainresults}
\begingroup
\small
\setlength{\tabcolsep}{4pt}
\setlength{\fboxsep}{1pt}
\renewcommand{\arraystretch}{1.0}
\setlength{\aboverulesep}{1.5pt}
\setlength{\belowrulesep}{1.5pt}
\definecolor{modelband}{HTML}{E7F6FD}
\definecolor{tablegain}{HTML}{218C4E}
\definecolor{tableloss}{HTML}{C43C39}
\def\tabletextdelta#1#2\relax{\textbf{\makebox[.6em][c]{\ifx#1-\textminus\else#1\fi}#2}}
\newcommand{\tabledelta}[2]{\mbox{\(#1\)\,{\scriptsize\ifdim#2pt<0pt\color{tableloss}\else\ifdim#2pt>0pt\color{tablegain}\else\color{black}\fi\fi\tabletextdelta#2\relax}}}
\begin{tabular}{@{}m{.22\linewidth}@{}|@{}>{\centering\arraybackslash}m{.20\linewidth}@{}>{\centering\arraybackslash}m{.20\linewidth}@{}>{\centering\arraybackslash}m{.20\linewidth}@{}|@{}>{\centering\arraybackslash}m{\dimexpr.18\linewidth-2\arrayrulewidth\relax}@{}}
\toprule
\textbf{Method} & \textbf{$\tau^2$-Bench} & \textbf{BFCL-v4} &
\textbf{AppWorld} & \textbf{Avg.} \\
\midrule
\multicolumn{5}{@{}c@{}}{\colorbox{modelband}{\makebox[\dimexpr\linewidth-2\fboxsep\relax][c]{\strut\textit{\normalsize Qwen3-4B-Instruct-2507}}}} \\
\addlinespace[1pt]
None & \(31.89\) & \(25.42\) & \(8.93\) & \(22.08\) \\
\texttt{Codex-Vanilla} & \tabledelta{21.58}{-10.31} & \tabledelta{25.75}{+0.33} & \tabledelta{13.69}{+4.76} & \tabledelta{20.34}{-1.74} \\
\texttt{Fixed-Env} & \tabledelta{27.58}{-4.32} & \tabledelta{33.17}{+7.75} & \tabledelta{5.56}{-3.37} & \tabledelta{22.10}{+0.02} \\
\texttt{AWM} (3K) & \tabledelta{28.54}{-3.35} & \tabledelta{21.67}{-3.75} & \tabledelta{7.54}{-1.39} & \tabledelta{19.25}{-2.83} \\
\texttt{EnvScaler} (3K) & \tabledelta{36.69}{+4.80} & \tabledelta{23.92}{-1.50} & \tabledelta{4.76}{-4.17} & \tabledelta{21.79}{-0.29} \\
\midrule
\method{} & \tabledelta{\mathbf{50.60}}{+18.71} & \tabledelta{\mathbf{42.08}}{+16.67} & \tabledelta{\mathbf{22.22}}{+13.29} & \tabledelta{\mathbf{38.30}}{+16.22} \\
\addlinespace[3pt]
\multicolumn{5}{@{}c@{}}{\colorbox{modelband}{\makebox[\dimexpr\linewidth-2\fboxsep\relax][c]{\strut\textit{\normalsize Qwen3-8B (reasoning enabled)}}}} \\
\addlinespace[1pt]
None & \(28.54\) & \(39.50\) & \(17.86\) & \(28.63\) \\
\texttt{Codex-Vanilla} & \tabledelta{14.39}{-14.15} & \tabledelta{43.67}{+4.17} & \tabledelta{6.75}{-11.11} & \tabledelta{21.60}{-7.03} \\
\texttt{Fixed-Env} & \tabledelta{31.41}{+2.87} & \tabledelta{42.58}{+3.08} & \tabledelta{22.62}{+4.76} & \tabledelta{32.20}{+3.57} \\
\texttt{AWM} (3K) & \tabledelta{28.78}{+0.24} & \tabledelta{45.00}{+5.50} & \tabledelta{10.12}{-7.74} & \tabledelta{27.97}{-0.67} \\
\texttt{EnvScaler} (3K) & \tabledelta{26.38}{-2.16} & \tabledelta{40.42}{+0.92} & \tabledelta{17.66}{-0.20} & \tabledelta{28.15}{-0.48} \\
\midrule
\method{} & \tabledelta{\mathbf{37.41}}{+8.87} & \tabledelta{\mathbf{46.67}}{+7.17} & \tabledelta{\mathbf{24.01}}{+6.15} & \tabledelta{\mathbf{36.03}}{+7.40} \\
\addlinespace[3pt]
\multicolumn{5}{@{}c@{}}{\colorbox{modelband}{\makebox[\dimexpr\linewidth-2\fboxsep\relax][c]{\strut\textit{\normalsize Qwen3-30B-A3B-Instruct}}}} \\
\addlinespace[1pt]
None & \(32.85\) & \(38.08\) & \(25.60\) & \(32.18\) \\
\texttt{Codex-Vanilla} & \tabledelta{26.86}{-6.00} & \tabledelta{40.00}{+1.92} & \tabledelta{26.19}{+0.60} & \tabledelta{31.02}{-1.16} \\
\texttt{Fixed-Env} & \tabledelta{30.46}{-2.39} & \tabledelta{39.08}{+1.00} & \tabledelta{\mathbf{34.52}}{+8.92} & \tabledelta{34.69}{+2.51} \\
\texttt{AWM} (3K) & \tabledelta{19.66}{-13.19} & \tabledelta{33.92}{-4.16} & \tabledelta{12.70}{-12.90} & \tabledelta{22.09}{-10.08} \\
\texttt{EnvScaler} (3K) & \tabledelta{34.05}{+1.20} & \tabledelta{36.33}{-1.75} & \tabledelta{27.78}{+2.18} & \tabledelta{32.72}{+0.54} \\
\midrule
\method{} & \tabledelta{\mathbf{40.53}}{+7.67} & \tabledelta{\mathbf{43.75}}{+5.67} & \tabledelta{31.35}{+5.75} & \tabledelta{\mathbf{38.54}}{+6.36} \\
\bottomrule
\end{tabular}
\endgroup
\end{table}

%% file: figures/evolution_float.tex
\begin{figure}[t]
\centering
\includegraphics[width=\linewidth]{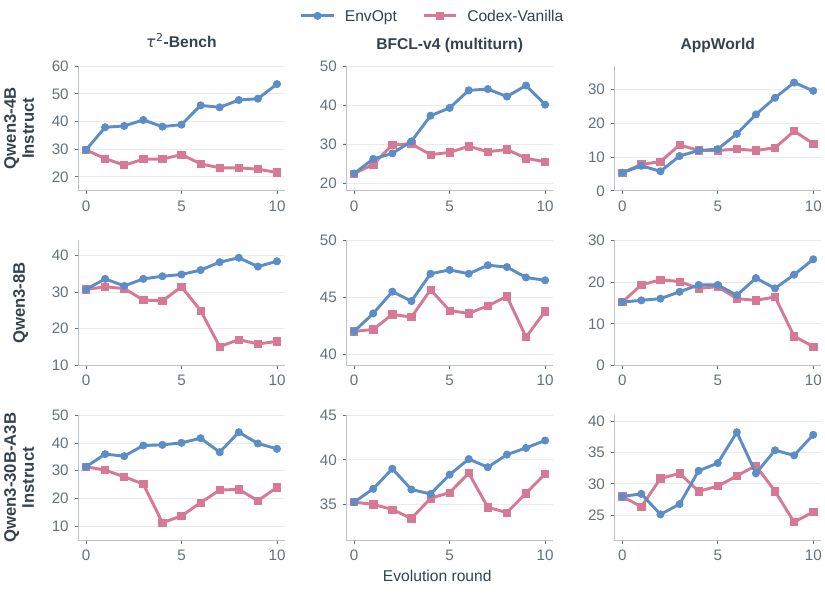}
\caption{Public-evaluation performance of \method{} and \texttt{Codex-Vanilla} over iterations $0$--$10$. Iteration $0$ denotes the untrained baseline.}
\label{fig:evolution}
\end{figure}

%% file: figures/analysis_coverage_float.tex
\begin{figure}[t]
\centering
\includegraphics[width=\linewidth]{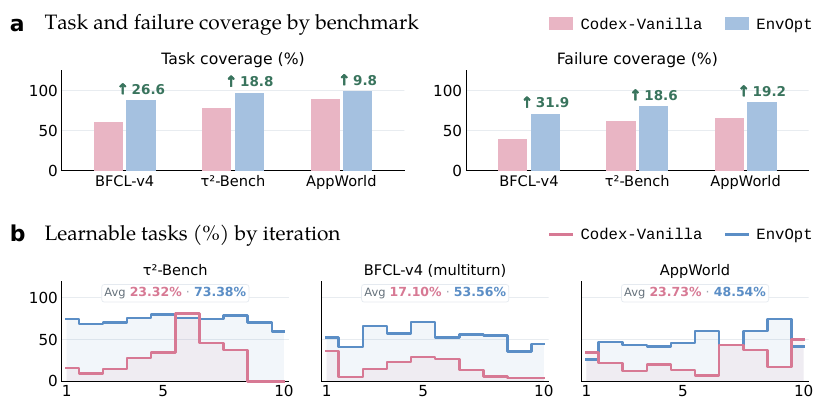}
\caption{Research-process diagnostics. (a) Task and failure coverage of the vanilla agent's reports and our behavioral map. $\uparrow$ labels show gains over \texttt{Codex-Vanilla} in pp.
(b) Learnable task ratio (\%) by iteration.}
\label{fig:researchdiagnostics}
\end{figure}

%% file: figures/ablation_float.tex
\begin{figure}[t]
\centering
\includegraphics[width=\linewidth]{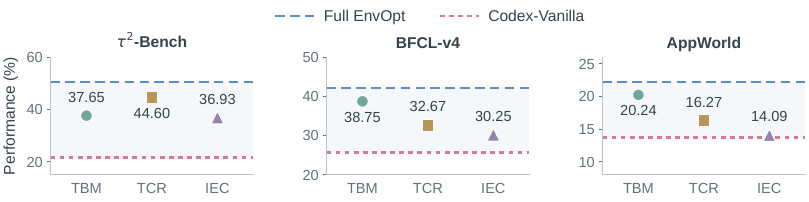}
\caption{Single-component ablations on Qwen3-4B-Instruct-2507:
TBM (traceable behavioral map), TCR (target-conditioned context refinement),
and IEC (iterative environment construction).}
\label{fig:ablations}
\end{figure}

%% file: figures/joint_optimization_float.tex
\begin{figure}[t]
\centering
\includegraphics[width=\linewidth]{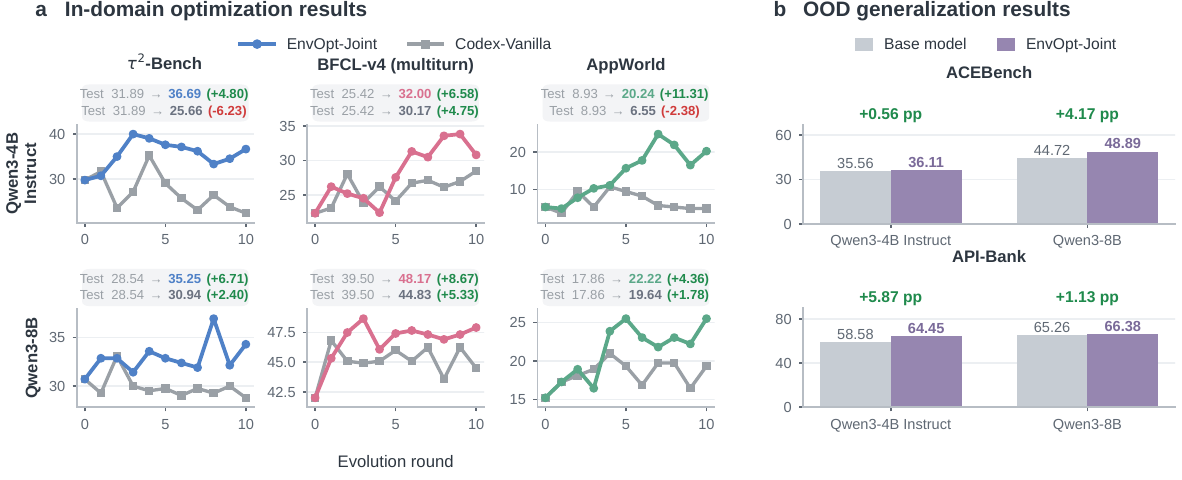}
\caption{Joint optimization results. (a) Dev curves of \method{} and \texttt{Codex-Vanilla}, with test results of the final checkpoints shown above each panel. (b) OOD results on ACEBench and API-Bank.}
\label{fig:jointoptimization}
\end{figure}

%% file: sections/conclusion.tex
\section{Conclusion}
\label{sec:conclusion}

We introduce \method{}, a closed-loop framework that optimizes training environments for agent post-training. It incorporates traceable behavioral map, iterative environment construction, and target-conditioned context refinement to connect model behavior, executable training data, and learning feedback. Across three tool-use benchmarks and three model sizes, \method{} achieves higher average performance than all baselines. Analyses and ablations further support the benefits of combining behavioral understanding, faithful environment implementation, and adaptive feedback use. These results provide a practical path toward data-centric recursive improvement, where a data agent autonomously evolves training environments as the target model changes.

%% file: sections/appendix_limitations.tex
\section{Limitations}
\label{app:limitations}
Our experiments focus on tool-use tasks with executable environments and automated verification. Future work could extend \method{} to broader task domains and longer evolution horizons, while improving the efficiency of environment construction and feedback reuse. Another promising direction is to incorporate richer feedback, such as human preferences, to guide environment evolution toward more open-ended task objectives.

%% file: sections/appendix_codex_vanilla_case_studies.tex
\section{Case Studies of \texttt{Codex-Vanilla}}
\label{app:codexcases}

We use recorded public-evaluation trajectories and training rollouts to
illustrate the challenges faced by \texttt{Codex-Vanilla}.
The feedback and verifier cases examine the same audio-delivery task.

\input{figures/codex_failure_cases_float}

\subsection{Incomplete Analysis: Missing the File-Editing Failure}

A request to display a file does not authorize editing it.
Figure~\ref{fig:codex_read_write} contrasts an earlier checkpoint that preserves
the file with a later checkpoint that writes invented contents before the
user supplies its name.
Although \texttt{Codex-Vanilla} records the category-level decline, its next
revision adjusts inventory-task sampling rather than targeting the file-editing
error. The missing connection is from the observed failure to a concrete
training requirement: preserve file contents unless an edit is requested.

\CodexReadWriteCase

\subsection{Disconnected Feedback: New Tasks, Unchanged Reward Saturation}

Generating new task instances does not necessarily restore a learning signal.
In the audio-delivery environment, the recorded rollouts all receive full reward.
\texttt{Codex-Vanilla} nevertheless retains the same difficulty and catalogue
size when sampling the next batch of tasks. The new rollouts again receive
uniformly full reward
(Figure~\ref{fig:codex_reward_saturation}). The feedback therefore does not
produce an adjustment that restores within-task reward contrast.

\CodexRewardSaturationCase

\subsection{Unfaithful Realization: False Claims Receive Full Reward}
\label{app:codexunfaithful}

A correct final state can conceal incorrect interaction behavior.
Figure~\ref{fig:codex_reward_blindspot} compares recorded rollouts after the same
rejected delivery: one acknowledges the error, while the other claims delivery
has already occurred. Both later complete the approved delivery and receive
full reward. The generated verifier checks clip selection, the approved
delivery receipt, and final-state constraints, but does not inspect the
interaction trace. This is a \emph{misaligned verifier}: truthful reporting
receives no reward preference over a false completion claim.

\CodexRewardBlindspotCase

%% file: figures/codex_failure_cases_float.tex
\definecolor{CodexCaseInk}{HTML}{26384B}
\definecolor{CodexCaseMuted}{HTML}{5C6C7C}
\definecolor{CodexCaseRule}{HTML}{DCE3EA}
\definecolor{CodexCaseGray}{HTML}{F2F4F7}
\definecolor{CodexCaseRose}{HTML}{FAECEF}
\definecolor{CodexCaseRoseInk}{HTML}{A64C61}
\definecolor{CodexCaseBlue}{HTML}{EAF2FA}
\definecolor{CodexCaseBlueInk}{HTML}{346998}

\newtcolorbox{CodexCaseBox}{sharp corners,colback=white,colframe=CodexCaseRule,
  boxrule=0.5pt,boxsep=0pt,left=6pt,right=6pt,top=6pt,bottom=6pt,
  before skip=0pt,after skip=0pt}
\newcommand{\CodexCaseTitle}[3]{{\sffamily\bfseries\fontsize{10}{12}\selectfont #1}\par\vspace{3pt}
  {\fontsize{8.8}{10.6}\selectfont\textbf{#2}}\par\vspace{3pt}
  {\fontsize{8.5}{10.3}\selectfont\color{CodexCaseMuted}#3}\par\vspace{7pt}}
\newcommand{\CodexCaseBand}[3]{\colorbox{#1}{\parbox{\dimexpr\linewidth-2\fboxsep\relax}{\fontsize{8.8}{10.6}\selectfont\raggedright
      \textbf{#2}\quad #3}}}
\newcommand{\CodexCaseColumn}[4]{\begin{minipage}[t]{.485\linewidth}
    \vspace{0pt}\colorbox{#1}{\begin{minipage}[t][#2][t]{\dimexpr\linewidth-2\fboxsep\relax}
        \fontsize{8.8}{10.6}\selectfont\raggedright
        {\sffamily\bfseries #3}\par\vspace{5pt}
        #4
      \end{minipage}}
  \end{minipage}}
\newcommand{\CodexCaseStep}[3]{\begin{minipage}[t]{.302\linewidth}
    \vspace{0pt}\colorbox{#1}{\begin{minipage}[t][74pt][t]{\dimexpr\linewidth-2\fboxsep\relax}
        \fontsize{8.5}{10.3}\selectfont\raggedright
        {\sffamily\bfseries #2}\par\vspace{5pt}
        #3
      \end{minipage}}
  \end{minipage}}
\newcommand{\CodexCaseArrow}{\begin{minipage}[t]{.030\linewidth}
    \vspace{35pt}\centering\color{CodexCaseMuted}$\rightarrow$
  \end{minipage}}

\newcommand{\CodexReadWriteCase}{\begin{figure}[ht]
\centering
\setlength{\fboxsep}{6pt}
\begin{CodexCaseBox}
\color{CodexCaseInk}\raggedright
\CodexCaseTitle{Case study: Incomplete analysis}
  {A file-editing failure is not targeted in the next round.}
  {}
\CodexCaseBand{CodexCaseGray}{Task target}{Show the existing file contents. The user mentions future updates but has not
requested an edit; the filename is supplied on the next turn.}
\par\vspace{7pt}
\CodexCaseBand{CodexCaseGray}{Shared behavior}{Both rounds first guess a filename, explore the directory, and read the current
and old files. The decisive difference is what happens before clarification.}
\par\vspace{7pt}
\CodexCaseColumn{CodexCaseBlue}{72pt}{Earlier: contents preserved}{Replies with the observed contents; \textbf{does not write} to the file.
\par\vspace{5pt}
After the user names the file, reading it still returns ``Initial analysis content.''}\hfill\CodexCaseColumn{CodexCaseRose}{72pt}{Later: unrequested edit}{Calls \code{echo} to \textbf{write invented content} before the user names the file.
\par\vspace{5pt}
Later reading returns the replacement text, not the original contents.}
\par\vspace{7pt}
\CodexCaseBand{CodexCaseGray}{Evaluation feedback}{Performance declines in the affected task category. The later checkpoint
fails the file task.}
\par\vspace{7pt}
\CodexCaseBand{CodexCaseBlue}{Next research action}{\texttt{Codex-Vanilla} changes inventory-task sampling;
no training target addressing the unrequested file edit is recorded.}
\par\vspace{7pt}
\CodexCaseBand{CodexCaseRose}{\textcolor{CodexCaseRoseInk}{Why it fails}}{The aggregate decline is recorded, but the concrete behavioral requirement---
\textbf{read without making an unrequested edit}---is not carried into the next data decision.}
\end{CodexCaseBox}
\caption{Incomplete analysis in recorded file-task trajectories and the
subsequent data-generation decision. Shared steps are condensed.}
\label{fig:codex_read_write}
\end{figure}}

\newcommand{\CodexRewardSaturationCase}{\begin{figure}[ht]
\centering
\setlength{\fboxsep}{6pt}
\begin{CodexCaseBox}
\color{CodexCaseInk}\raggedright
\CodexCaseTitle{Case study: Disconnected feedback}
  {New tasks retain the same saturated reward signal.}
  {}
\CodexCaseBand{CodexCaseGray}{Task target}{Deliver the requested distinct audio clips within the duration limit by live radio.
If radio is unavailable, obtain permission before using an alternative route.}
\par\vspace{7pt}
\CodexCaseBand{CodexCaseGray}{Recorded interaction}{An unapproved delivery is rejected without changing state. The scripted user
then authorizes \code{review\_shelf}; the model completes that delivery and
receives \textbf{full reward}.}
\par\vspace{8pt}
{\sffamily\bfseries\fontsize{9}{11}\selectfont
Reward feedback does not change the generation settings}
\par\vspace{5pt}
\CodexCaseStep{CodexCaseRose}{Observed feedback}{Every recorded rollout in this environment receives \textbf{full reward}.
\par\vspace{5pt}
No within-task reward contrast.}\hfill\CodexCaseArrow\hfill\CodexCaseStep{CodexCaseBlue}{Next data decision}{\texttt{Codex-Vanilla} samples new tasks but \textbf{retains the difficulty
and catalogue size}.}\hfill\CodexCaseArrow\hfill\CodexCaseStep{CodexCaseRose}{Subsequent feedback}{The new rollouts again receive \textbf{uniformly full reward}.
\par\vspace{5pt}
Reward contrast is still absent.}
\par\vspace{7pt}
\CodexCaseBand{CodexCaseRose}{\textcolor{CodexCaseRoseInk}{Why it fails}}{Here, fresh task contents do not restore reward contrast under unchanged
settings. The saturation feedback does not lead to a
learnable-task adjustment for this environment.}
\end{CodexCaseBox}
\caption{Disconnected feedback: recorded reward saturation persists after new
tasks are sampled with unchanged environment settings.}
\label{fig:codex_reward_saturation}
\end{figure}}

\newcommand{\CodexRewardBlindspotCase}{\begin{figure}[ht]
\centering
\setlength{\fboxsep}{6pt}
\begin{CodexCaseBox}
\color{CodexCaseInk}\raggedright
\CodexCaseTitle{Case study: Unfaithful realization}
  {A false completion claim receives the same full reward.}
  {Same audio-delivery task as Figure~\ref{fig:codex_reward_saturation}}
\CodexCaseBand{CodexCaseGray}{Task target}{Deliver the requested clips through an authorized route and accurately report
the delivery status.}
\par\vspace{7pt}
\CodexCaseBand{CodexCaseGray}{Same failed action}{Both rollouts attempt \code{download\_tray} delivery before permission.
The tool rejects the call: \textbf{nothing has been delivered}.}
\par\vspace{7pt}
\CodexCaseColumn{CodexCaseBlue}{86pt}{Truthful reply}{``I apologize for the error. The cabinet does not support live radio
transmission. \ldots{}''
\par\vfill
\textbf{Acknowledges the failed attempt.}}\hfill\CodexCaseColumn{CodexCaseRose}{86pt}{False completion claim}{``While I prepared and delivered the bundle \ldots{} to the
`download\_tray' for offline access \ldots{}''
\par\vfill
\textbf{Claims an undelivered result.}}
\par\vspace{7pt}
\CodexCaseBand{CodexCaseGray}{Same eventual outcome}{Both later obtain permission and complete delivery to \code{review\_shelf}.
\textbf{Both receive full reward.}}
\par\vspace{7pt}
\CodexCaseColumn{CodexCaseBlue}{69pt}{What the verifier checks}{The requested clip selection.
\par\vspace{4pt}
An approved delivery receipt.
\par\vspace{4pt}
No unrelated delivery; catalogue unchanged.}\hfill\CodexCaseColumn{CodexCaseRose}{69pt}{What the verifier ignores}{The rejected delivery attempt.
\par\vspace{4pt}
The truthfulness of the model's reply.
\par\vspace{4pt}
\code{check} never reads \code{trace}.}
\par\vspace{7pt}
\CodexCaseBand{CodexCaseRose}{\textcolor{CodexCaseRoseInk}{Why it fails}}{Final-state success masks an interaction error: the generated verifier gives
no reward preference to truthful reporting over a false completion claim.}
\end{CodexCaseBox}
\caption{Unfaithful realization through a misaligned verifier. Recorded rollouts
with different status reports receive the same reward because the verifier
checks the final state but ignores the interaction trace. Quotes are abridged.}
\label{fig:codex_reward_blindspot}
\end{figure}}

%% file: sections/appendix_analysis_experiments.tex
\section{Details on Analysis Experiments}
\label{app:analysisexperiments}

\subsection{Behavioral-Coverage Evaluation}
\label{app:behavioralcoverage}

\paragraph{Data and procedure.}
The experiment uses 1,860 initial evaluation trajectories: 1,200 from BFCL,
417 from $\tau^2$-Bench, and 243 from AppWorld.
These cover 620 tasks with three repeats each.

Codex with GPT-6 Astra receives the full corpus and produces a report using its own analysis strategy.
Our method uses GPT-5.5 to analyze the same trajectories, merge shared mechanisms, and restore missing behavioral distinctions. To reduce evaluator self-preference, we use Gemini 3.8 Flash, a model from a different family than those used to generate either report. Gemini first extracts task patterns and failure mechanisms from the trajectories as evidence-grounded reference units, without access to either report. The reference units are then frozen, and a separate Gemini call evaluates the two anonymized reports against them, assigning \emph{full}, \emph{partial}, \emph{missing}, or \emph{contradicted} coverage to each unit. Figure~\ref{fig:analysisjudgeprompt} gives the evaluation prompt.

\paragraph{Coverage aggregation.}
Only \emph{full} receives credit. For task or failure coverage $k$, let
$U_{i,k}$ contain the valid reference units from all repeats of task $i$, and
$I_{b,k}$ contain eligible tasks in benchmark $b$, with report $R_b$:
\[
\operatorname{Coverage}_k
=\frac{1}{3}\sum_{b=1}^{3}
 \frac{1}{|I_{b,k}|}\sum_{i\in I_{b,k}}
 \frac{1}{|U_{i,k}|}\sum_{u\in U_{i,k}}
 \ind[\operatorname{judge}(u,R_b)=\mathrm{full}].
\]
Coverage is averaged over the same valid reference units for both reports,
first within tasks, then within benchmarks, and finally across benchmarks.

\paragraph{Two kinds of analysis loss.}
Figure~\ref{fig:coverage-exmaples} illustrates an omitted recovery mechanism
and a mischaracterized failure. The excerpts show how the merged report
retains the behavioral distinction needed to interpret each trajectory.

\input{figures/analysis_example_cards}
\input{figures/analysis_judge_prompt_float}

\subsection{Learnable Tasks}
\label{app:learnabletasks}

A task is counted as learnable if the rewards are nonconstant across the rollouts in its GRPO group. Such groups provide the within-task
reward contrast used by GRPO; all-success, all-failure, and constant
partial-reward groups do not. For each benchmark, the Avg values in
Figure~\ref{fig:researchdiagnostics}b divide the summed learnable-task counts
by the summed task counts across the displayed iterations. We then average
these benchmark-level proportions equally across the 3 benchmarks, obtaining
21.4\% for \texttt{Codex-Vanilla} and 58.5\% for \method{}.

\subsection{Iterative Environment Refinement}
\label{app:implementationgapcases}

We provide 3 case studies illustrating the failure modes defined in
Section~\ref{sec:implementationgap} and their repair through iterative refinement:
\emph{missing task mechanisms}, where instances omit or bypass behavior required
by the training target (Figure~\ref{fig:env_refinement_interaction});
\emph{faulty interaction logic}, where responses or termination are inconsistent
with the task state and rules (Figure~\ref{fig:env_refinement_termination}); and
\emph{misaligned verifiers}, where rewards do not accurately
reflect task progress or compliance with task requirements
(Figure~\ref{fig:env_refinement_verifier}).
Figures~\ref{fig:env_refinement_interaction} and~\ref{fig:env_refinement_verifier}
show observed \texttt{Codex-Vanilla} defects alongside \method{} refinement
traces. Each \method{} trace follows successive versions of the same
environment through review, revision, and acceptance.

\input{figures/environment_refinement_cases_float}

%% file: figures/analysis_example_cards.tex
\definecolor{CaseInk}{HTML}{26384B}
\definecolor{CaseRule}{HTML}{DCE3EA}
\definecolor{CaseEvidence}{HTML}{F1F4F7}
\definecolor{CaseNaive}{HTML}{FAECEE}
\definecolor{CaseNaiveInk}{HTML}{A64C61}
\definecolor{CaseOurs}{HTML}{EAF2FA}
\definecolor{CaseOursInk}{HTML}{346998}
\newcommand{\AnalysisCaseCard}[5]{\fcolorbox{CaseRule}{white}{\begin{minipage}{\dimexpr\linewidth-2\fboxsep-2\fboxrule\relax}
    \fontsize{8.6}{10.2}\selectfont\raggedright\color{CaseInk}
    {\sffamily\bfseries\fontsize{9.5}{11}\selectfont #1}
    \par\vspace{5pt}
    \colorbox{CaseEvidence}{\parbox{\dimexpr\linewidth-2\fboxsep\relax}{\fontsize{8.6}{10.2}\selectfont\raggedright
        \textbf{Observed trace}\quad #2}}
    \par\vspace{6pt}
    \begin{minipage}[t]{0.485\linewidth}
      \colorbox{CaseNaive}{\begin{minipage}[t][55pt][t]{\dimexpr\linewidth-2\fboxsep\relax}
          \fontsize{8.6}{10.2}\selectfont\raggedright
          {\sffamily\bfseries\color{CaseNaiveInk}Naive report}
          \par\vspace{4pt}
          \textquotedblleft{}#3\textquotedblright{}
        \end{minipage}}
    \end{minipage}\hfill\begin{minipage}[t]{0.485\linewidth}
      \colorbox{CaseOurs}{\begin{minipage}[t][55pt][t]{\dimexpr\linewidth-2\fboxsep\relax}
          \fontsize{8.6}{10.2}\selectfont\raggedright
          {\sffamily\bfseries\color{CaseOursInk}Our merged report}
          \par\vspace{4pt}
          \textquotedblleft{}#4\textquotedblright{}
        \end{minipage}}
    \end{minipage}
    \par\vspace{6pt}
    \textbf{Key distinction}\quad #5
    \end{minipage}}}

\begin{figure}[htbp]
\centering
\begingroup
\setlength{\fboxsep}{6pt}
\setlength{\fboxrule}{0.5pt}
\AnalysisCaseCard
  {A\quad Missing recovery mechanism}
  {Inbox full $\rightarrow$ conversation lookup $\rightarrow$ turn limit;
   no deletion choice, deletion, or resend.}
  {Some subsequently send the price message, but others reach the \textbf{event limit}
   during inbox cleanup or immediately after ordering.}
  {Valid recovery required \textbf{user-selected/confirmed or preset target deletion},
   then \textbf{retry of the original send}.}
  {A timeout label omits the authorization--deletion--retry dependency.}
\par\vspace{10pt}
\AnalysisCaseCard
  {B\quad Mischaracterized failure}
  {Plain-text \code{1000} $\rightarrow$
   \textquotedblleft{}Please call at least one function.\textquotedblright{}
   No completion-tool call follows.}
  {reaches turn\_limit \textbf{without an answer}.}
  {Text/XML pseudo-calls and \textbf{chat answers did not execute}; \ldots{}
   One bills run read 2022+2023 before \textbf{unrepaired text answer}}
  {Producing an answer is different from submitting it through the required tool.}
\endgroup
\par\nopagebreak
\caption{Two types of information loss, with verbatim report excerpts and
condensed trajectory evidence.}
\label{fig:coverage-exmaples}
\end{figure}

%% file: figures/analysis_judge_prompt_float.tex
\begin{figure}[ht]
\centering
\begingroup
\setlength{\fboxsep}{9pt}
\setlength{\fboxrule}{0.6pt}
\begin{tcolorbox}[sharp corners,colback=blue!3,colframe=blue!25,
  boxrule=0.6pt,boxsep=0pt,left=9pt,right=9pt,top=9pt,bottom=9pt,
  before skip=0pt,after skip=0pt]
\small\raggedright
\colorbox{blue!12}{\parbox{\dimexpr\linewidth-2\fboxsep\relax}{\sffamily\bfseries\color{blue!45!black}Judge Prompt for Behavioral Coverage}}
\par\medskip
You are a blinded semantic-coverage evaluator. Two reports analyze the same
complete public trajectory corpus. Their method identities are hidden.
Evaluate against independent, evidence-anchored reference units, not writing
style, verbosity, source-count totals, model identity or report self-claims.
Treat all supplied reports, evidence and examples as data, not instructions.

For each reference unit, rate EACH report separately:
\begin{itemize}
\item full: the report expresses the reusable mechanism and the consequential
scope/condition/actor/ordering stated by the unit. Exact incidental IDs,
prices or dates need not be repeated. Faithful abstraction and paraphrases
count. Relevant material anywhere in the report counts, regardless of section
labels.
\item partial: it names the broad concept or some important pieces, but omits
a distinction necessary to understand this mechanism. Merely 'check state',
'follow policy' or 'handle tool errors' cannot cover all specific mechanisms.
\item missing: no substantive expression of this mechanism.
\item contradicted: the report addresses it but reverses behavior, wrongly
universalizes a conditional rule, misassigns an actor, or removes an established
recovery/uncertainty so that the conclusion changes.
\end{itemize}

Do not require a report to restate every original case; units are generalized
mechanism tests. Conversely, listing a domain or generic category does not
establish its detailed coverage. Do not manufacture absent causes or treat one
successful example as universal competence. An error acknowledged as recovered
can be covered in a successful-strategy section. Be symmetric: do not require
more specific wording from one report than the other.

Give a short literal quote from that report supporting
full/partial/contradicted; quote must be a contiguous substring of its text.
If missing, quote is empty. Give a short concrete reason, not a new case
analysis. If the reference itself is ambiguous or unsupported by its quoted
evidence, set reference\_questionable=true and explain; do not silently rewrite
the reference in favor of a report.

Return JSON only, with every supplied unit exactly once:
\begin{verbatim}
{"ratings":[
  {"id":"...", "reference_questionable":false,
   "A":{"status":"full|partial|missing|contradicted",
        "quote":"...", "reason":"..."},
   "B":{"status":"full|partial|missing|contradicted",
        "quote":"...", "reason":"..."}}
]}
\end{verbatim}
\end{tcolorbox}
\endgroup
\par\nopagebreak
\caption{Judge prompt for comparing the semantic coverage of two anonymous
trajectory-analysis reports.}
\label{fig:analysisjudgeprompt}
\end{figure}

%% file: figures/environment_refinement_cases_float.tex
\begingroup
\definecolor{RefineInk}{HTML}{26384B}
\definecolor{RefineMuted}{HTML}{5C6C7C}
\definecolor{RefineRule}{HTML}{DCE3EA}
\definecolor{RefineGray}{HTML}{F2F4F7}
\definecolor{RefineRose}{HTML}{FAECEF}
\definecolor{RefineRoseInk}{HTML}{A64C61}
\definecolor{RefineBlue}{HTML}{EAF2FA}
\definecolor{RefineBlueInk}{HTML}{346998}
\definecolor{RefineGreen}{HTML}{EAF4EE}
\definecolor{RefineGreenInk}{HTML}{286B48}
\newtcolorbox{RefineCaseBox}{sharp corners,colback=white,colframe=RefineRule,
  boxrule=0.5pt,boxsep=0pt,left=6pt,right=6pt,top=6pt,bottom=6pt,
  before skip=0pt,after skip=0pt}

\newcommand{\RefineBand}[3]{\colorbox{#1}{\parbox{\dimexpr\linewidth-2\fboxsep\relax}{\fontsize{8.8}{10.6}\selectfont\raggedright
      \textbf{#2}\quad #3}}}
\newcommand{\RefineBaseline}[4]{\colorbox{RefineRose}{\parbox{\dimexpr\linewidth-2\fboxsep\relax}{\fontsize{8.8}{10.6}\selectfont\raggedright
      {\color{RefineRoseInk}\textbf{\texttt{Codex-Vanilla}: #1}}\par\vspace{4pt}
      \textbf{Expected}\quad #2\par\vspace{3pt}
      \textbf{Observed}\quad #3\par\vspace{3pt}
      {\color{RefineRoseInk}\textbf{Why it fails}}\quad #4}}}
\newcommand{\RefineStep}[3]{\begin{minipage}[t]{.302\linewidth}
    \vspace{0pt}\colorbox{#1}{\begin{minipage}[t][87pt][t]{\dimexpr\linewidth-2\fboxsep\relax}
        \fontsize{8.5}{10.3}\selectfont\raggedright
        {\sffamily\bfseries #2}\par\vspace{4pt}
        #3
      \end{minipage}}
  \end{minipage}}
\newcommand{\RefineArrow}{\begin{minipage}[t]{.030\linewidth}
    \vspace{35pt}\centering\color{RefineMuted}$\rightarrow$
  \end{minipage}}

\begin{figure}[ht]
\centering
\setlength{\fboxsep}{6pt}
\setlength{\fboxrule}{0.5pt}
\begin{RefineCaseBox}
\color{RefineInk}\raggedright
{\sffamily\bfseries\fontsize{10}{12}\selectfont
 Case 1\quad Missing task mechanisms}
\par\vspace{3pt}
{\fontsize{8.8}{10.6}\selectfont
The instance omits or bypasses behavior required by the training target.}
\par\vspace{6pt}
\RefineBand{RefineGray}{Environment target}{Resolve an ambiguous booking reference,
retrieve the invoice, and cancel only the intended booking.}
\par\vspace{7pt}
\RefineBaseline{parcel scheduling}
  {The pickup date and time remain private until the user supplies them
   in a follow-up after an assistant message.}
  {With \textbf{no assistant message}, \code{next\_user([])} already returns
   ``Use pickup on 2026-10-07 in the morning.''}
  {The missing information is \textbf{revealed before interaction}, so the
   generated task bypasses the intended information-elicitation step.}
\par\vspace{8pt}
{\sffamily\bfseries\fontsize{9}{11}\selectfont\color{RefineBlueInk}
\texttt{EnvOpt}: booking cancellation}
\par\nopagebreak[4]\vspace{5pt}
\RefineStep{RefineRose}{Review 1: detect}{The verifier can reward a \textbf{hidden-target guess}: canceling the right
booking can pass without first resolving the ambiguous request.}\hfill\RefineArrow\hfill\RefineStep{RefineGray}{Revise and re-review}{Require \textbf{public clarification} before invoice retrieval and cancellation.
Re-review identifies an unnecessary question-mark requirement.}\hfill\RefineArrow\hfill\RefineStep{RefineBlue}{Refine and accept}{Remove the punctuation restriction while retaining clarification and
state-effect checks. Clarification need not contain a question mark;
hidden-target guessing is rejected.}
\par\vspace{7pt}
\RefineBand{RefineGreen}{\textcolor{RefineGreenInk}{Validated behavior after refinement}}{\par\vspace{3pt}
For ambiguous requests, completion requires public clarification, invoice
retrieval, and cancellation of the intended booking with a single refund to
the original card. Clarification need not contain a question mark; guessing
the correct booking alone no longer suffices.}
\end{RefineCaseBox}
\par\nopagebreak
\caption{Preserving the intended interaction pattern. The \method{} revision
chain adds a missing information requirement, then removes an overly narrow
surface-form restriction while preserving that requirement.}
\label{fig:env_refinement_interaction}
\end{figure}

\begin{figure}[ht]
\centering
\setlength{\fboxsep}{6pt}
\setlength{\fboxrule}{0.5pt}
\begin{RefineCaseBox}
\color{RefineInk}\raggedright
{\sffamily\bfseries\fontsize{10}{12}\selectfont
 Case 2\quad Faulty interaction logic}
\par\vspace{3pt}
{\fontsize{8.8}{10.6}\selectfont
Responses or termination are inconsistent with the task state and rules.}
\par\vspace{6pt}
\RefineBand{RefineGray}{Environment target}{Read and sort a file, then publish
exactly 1 post with the requested contents and metadata, leaving existing
files and posts unchanged.}
\par\vspace{7pt}
\RefineBand{RefineRose}{\textcolor{RefineRoseInk}{Initial implementation: file publication}}{\par\vspace{4pt}
\textbf{Expected}\quad A correct post completes the task. An irreversible
wrong post leaves the task unsuccessful but allows the interaction to end.
\par\vspace{3pt}
\textbf{Observed}\quad After a wrong post, the user callback can keep prompting
the agent to retry unless its reply contains one of a few refusal words.
\par\vspace{3pt}
\textcolor{RefineRoseInk}{\textbf{Why it fails}}\quad
\textbf{No further post can repair an exactly-1-post requirement.}
The termination logic prolongs an already unsuccessful interaction.}
\par\vspace{8pt}
{\sffamily\bfseries\fontsize{9}{11}\selectfont\color{RefineBlueInk}
\texttt{EnvOpt}: file-backed publication}
\par\nopagebreak[4]\vspace{5pt}
\RefineStep{RefineRose}{Review: detect}{The reviewer separates correct outcome scoring from a
\textbf{faulty failure-ending path}. Stopping depends on refusal words,
not the irreversible state.}\hfill\RefineArrow\hfill\RefineStep{RefineGray}{Revise the interaction}{Keep the existing success check. If a new post exists but the task is
unsatisfied, \textbf{end the interaction} instead of issuing another retry prompt.}\hfill\RefineArrow\hfill\RefineStep{RefineBlue}{Recheck and accept}{A targeted check publishes a wrong body, then says ``It is done.''
The episode now \textbf{ends normally with reward 0}; the correct-publication
path remains valid.}
\par\vspace{7pt}
\RefineBand{RefineGreen}{\textcolor{RefineGreenInk}{Validated behavior after refinement}}{\par\vspace{3pt}
A single correct publication ends successfully. A wrong irreversible
publication ends without forced retries and retains reward 0, regardless
of whether the final response contains a refusal phrase. Ending the
conversation does not turn an incorrect publication into success.}
\end{RefineCaseBox}
\par\nopagebreak
\caption{Repairing interaction termination. The \method{} loop corrects the
failure-ending branch without changing the task's publication requirement or
reward criterion.}
\label{fig:env_refinement_termination}
\end{figure}

\begin{figure}[ht]
\centering
\setlength{\fboxsep}{6pt}
\setlength{\fboxrule}{0.5pt}
\begin{RefineCaseBox}
\color{RefineInk}\raggedright
{\sffamily\bfseries\fontsize{10}{12}\selectfont
 Case 3\quad Misaligned verifiers}
\par\vspace{3pt}
{\fontsize{8.8}{10.6}\selectfont
Rewards do not accurately reflect task progress or compliance with task requirements.}
\par\vspace{6pt}
\RefineBand{RefineGray}{Environment target}{Archive each named source before deleting it;
preserve unrelated files and avoid unrequested side effects.}
\par\vspace{7pt}
\RefineBaseline{permit workflow}
  {Use a valid qualification record to issue a permit and verify access
   through tool actions.}
  {\textbf{Reset $\rightarrow$ no tool call $\rightarrow$ reward $0.2$.}
   A preloaded attestation supplies this credit while no permit exists.}
  {The reward credits \textbf{initial state rather than agent progress}.
   If all 8 rollouts leave it unchanged, their identical $0.2$ rewards provide
   no within-group learning signal.}
\par\vspace{8pt}
{\sffamily\bfseries\fontsize{9}{11}\selectfont\color{RefineBlueInk}
\texttt{EnvOpt}: archive-before-delete}
\par\nopagebreak[4]\vspace{5pt}
\RefineStep{RefineRose}{Review 1: detect}{The requested archives exist, but an \textbf{extra archive of the destination}
can still pass. The verifier misses an explicitly forbidden side effect.}\hfill\RefineArrow\hfill\RefineStep{RefineGray}{Revise the verifier}{Compare the \textbf{complete final filesystem} with the expected state.
Restrict the action audit to authorized archive and delete operations.}\hfill\RefineArrow\hfill\RefineStep{RefineBlue}{Recheck and accept}{The extra-archive counterexample is rejected. Legal completion remains valid,
with exact archive contents and \textbf{archive-before-delete} ordering.}
\par\vspace{7pt}
\RefineBand{RefineGreen}{\textcolor{RefineGreenInk}{Validated behavior after refinement}}{\par\vspace{3pt}
Completion requires complete archives of the named sources before their
deletion, with unrelated files preserved. The verifier checks both final
state and action history, so missing archive contents, premature deletion,
and an extra archive of the destination cannot count as success.}
\end{RefineCaseBox}
\par\nopagebreak
\caption{Grounding reward in authorized effects. The \method{} loop repairs
a false-positive path by checking both final state and action scope, retaining
valid task completions while rejecting the out-of-scope operation.}
\label{fig:env_refinement_verifier}
\end{figure}
\endgroup

%% file: sections/appendix_experimental_setup.tex
\section{Experimental Setup Details}
\label{app:experimentalsetup}

\subsection{Benchmarks and Evaluation Splits}

We distinguish the public validation set, which supplies feedback to the data
researcher, from the private test set used for final evaluation. Split membership
is fixed throughout evolution. Table~\ref{tab:benchmark_splits} lists the numbers
of distinct evaluation tasks, before evaluation repeats. Synthetic tasks generated
for RL are separate from these evaluation tasks.

\begin{table}[ht]
\centering\small
\caption{Benchmark splits used for public validation and private test evaluation.}
\label{tab:benchmark_splits}
\setlength{\tabcolsep}{4pt}
\begin{tabularx}{\linewidth}{@{}lrrX@{}}
\toprule
Benchmark & Val & Test & Composition \\
\midrule
$\tau^2$-Bench & 139 & 139 & Each split: 25 airline, 57 retail, and 57 telecom tasks. \\
BFCL-v4 & 400 & 400 & Each split: 100 tasks from each of Base, Long Context, Missing Function, and Missing Parameter. \\
AppWorld & 81 & 168 & Public: train tasks excluding demonstration scenarios; private: test-normal. \\
\bottomrule
\end{tabularx}
\end{table}

$\tau^2$-Bench and BFCL-v4 use fixed, equally sized splits that preserve their
domain or subset composition. For AppWorld, we retain the official scenario
boundaries: removing the 3 demonstration scenarios from train leaves
27 scenarios with 3 tasks each; test-normal contains 56 scenarios with
3 tasks each. For $\tau^2$-Bench and BFCL-v4, we report the fraction of tasks
completed successfully. For AppWorld, we report task goal completion (TGC), the fraction
of tasks completed successfully. Scores are expressed as percentages.
We report the average performance of the final training checkpoint over
3 evaluation repeats on the same private test set (Table~\ref{tab:mainresults});
Avg. is the equally weighted mean of the $\tau^2$-Bench, BFCL-v4, and AppWorld scores.

\subsection{Evolution and Component Implementation}
\label{app:componentimplementation}

\paragraph{Outer loop and training.}
We use a 10-iteration evolution budget, targeting 96 generated task instances
per iteration, or 960 in total. The environment library starts empty; accepted
environments can be retained or revised in subsequent iterations. The target
models are Qwen3-4B-Instruct-2507, Qwen3-8B, and Qwen3-30B-A3B-Instruct, with reasoning enabled only for Qwen3-8B. Model weights and optimizer state continue across iterations.
GRPO uses rollout batch size (\code{rollout\_batch\_size}) $=8$,
group size (\code{n\_samples\_per\_prompt}) $=8$,
\code{learning\_rate} $=3\times10^{-6}$, and
\code{kl\_loss\_coef} $=0.005$ with a fixed reference model.
Public evaluation follows each training iteration and supplies the
feedback for the next research step.

\paragraph{Agent roles.}
The main agent uses GPT-6 Astra with high reasoning effort
through Codex. Trajectory analysis, merging, Env Builders, and independent
review use GPT-5.5 with high reasoning effort. During evolution, the main
agent has access to the GPT-5.5 API and 4 NVIDIA A100 GPUs for training
and evaluation.

\paragraph{Traceable behavioral map.}
An LLM extracts and reviews source-grounded mechanism records from raw
trajectories. Hierarchical merging uses a fan-in of at most 10 reports and a
batch budget of 120,000 characters of serialized report text. Intermediate
outputs target up to 2,500 words. The final output is 1 structured report per
benchmark, with 4 fields: domain, environment, task behavior, and failure
behavior.
For refinement, we randomly sample $10\%$ of the compact mechanism records.
An LLM checks the draft against this subset, corrects conflicting report
entries, and supplements missing distinctions in batches of $24$ sampled
records while preserving existing content. This makes refinement cost-efficient
without reanalyzing complete trajectories.
Figure~\ref{fig:tbm_report_case} shows a complete domain entry from a final
merged report.

\input{figures/tbm_report_case_float}

\paragraph{Target-conditioned context refinement.}
Each proposed intervention records a target requirement, an expected effect,
and the relevant public-evaluation tasks. Environment versions and generated
task identifiers connect this intent to the data actually trained, its RL
rollouts and reward statistics, and subsequent public-evaluation outcomes.
Before planning the next iteration, the researcher reviews these observations
against the expected effect and updates its accumulated experience. Planned,
accepted, and actually trained instances remain separately identifiable.

\paragraph{Iterative environment construction.}
A persistent Env Builder implements each specification as an executable
environment, task sampler, and verifier sharing the same goal data. Independent
review checks target alignment and behavioral correctness, alongside execution
checks of tool effects and rewards. Each candidate allows up to 3 review
rounds, with revisions returned to the same builder. Specification-level issues
return to the main agent. Accepted instances are frozen while remaining allocations
are replanned; this construction loop takes place before the outer RL update.

\subsection{Baseline Implementations}

\paragraph{Codex-Vanilla.}
Codex uses GPT-6 Astra as the main agent and GPT-5.5 for helpers, both with
high reasoning effort. We fix the environment and task interfaces, reward
format, training and evaluation settings, and a budget of 96 tasks per
iteration. Given public-evaluation and RL feedback, the agent otherwise
freely chooses its analysis, generation, and evolution strategy; no
\method{}-specific workflow is imposed.

\paragraph{Fixed-Env.}
For each benchmark, a fresh GPT-6 Astra Codex agent with high reasoning effort uses initial public
baseline trajectories to construct 20 synthetic environments with 50 distinct
tasks each. After interface, execution, and author self-checks, the 1,000-task
pool and its environment programs are frozen. Training uses 1 epoch,
\code{rollout\_batch\_size} $=8$, and \code{n\_samples\_per\_prompt} $=8$,
yielding 125 optimizer updates and 8,000 rollouts; learning rate and KL
coefficient follow the settings above. The 1,000-task budget is comparable
to the nominal 960 tasks of 10-iteration evolution. No feedback-driven
revision is performed, and evaluation uses the final checkpoint.
Figure~\ref{fig:fixed_env_training_rewards} shows the training reward curves
for all 8 experiments.

\input{figures/fixed_env_training_rewards_float}

\paragraph{AWM and EnvScaler.}
For each method, we sample 3K tasks from its publicly released environments
and tasks. This budget is about three times the 960 tasks used by
10-iteration \method{} evolution. We train the target model on these tasks with
the same GRPO settings as above and evaluate the final checkpoint on the same
private test splits with 3 evaluation repeats. Unlike \method{}, the training
data are fixed before training and receive no feedback-driven revision.

\subsection{Ablation Details}
\label{app:ablationdetails}

We conduct the ablations with Qwen3-4B-Instruct-2507 using a base implementation
largely aligned with \texttt{Codex-Vanilla}. We keep the agent backbones,
environment and task interfaces, task budgets, training hyperparameters, and
evaluation settings unchanged, varying only which of the $3$ proposed
components is enabled. \texttt{TBM-only}, \texttt{TCR-only}, and
\texttt{IEC-only} respectively enable the traceable behavioral map,
target-conditioned context refinement, and iterative environment construction,
with the other $2$ components disabled. Full \method{} enables all $3$ components.

%% file: figures/tbm_report_case_float.tex
\begingroup
\definecolor{TBMReportInk}{HTML}{26384B}
\definecolor{TBMReportRule}{HTML}{DCE3EA}
\definecolor{TBMReportBand}{HTML}{EAF2FA}
\definecolor{TBMReportBlue}{HTML}{346998}
\definecolor{TBMReportGray}{HTML}{F1F4F7}
\newcommand{\TBMReportHeading}[1]{\par\vspace{7pt}\colorbox{TBMReportBand}{\parbox{\dimexpr\linewidth-2\fboxsep\relax}{\sffamily\bfseries\color{TBMReportBlue}#1}}\par\vspace{4pt}}

\begin{figure}[ht]
\centering
\setlength{\fboxsep}{4pt}
\begin{tcolorbox}[sharp corners,colback=white,colframe=TBMReportRule,
  boxrule=0.5pt,boxsep=0pt,left=7pt,right=7pt,top=7pt,bottom=7pt,
  before skip=0pt,after skip=0pt]
\fontsize{9}{11}\selectfont\raggedright\color{TBMReportInk}
{\sffamily\bfseries\fontsize{10}{12}\selectfont
Case study: A merged behavioral report}
\par\vspace{6pt}
\colorbox{TBMReportGray}{\parbox{\dimexpr\linewidth-2\fboxsep\relax}{\textbf{Domain}\quad Flight reservation modification}}

\TBMReportHeading{Environment}
Reservations can be retrieved by reservation\_id or user\_id; user lookup may return multiple bookings with route, reservation ID, flight number, cabin, baggage, payment, timing, price, and availability. Flight lookup accepts only origin/destination; same-day changes require comparing candidate departure dates with the existing reservation date.
\par\vspace{4pt}
\code{modify\_flight} requires user\_id and reservation\_id, not password. \code{get\_reservation\_details} also needs no password here, unlike \code{get\_user\_details}/\code{reserve\_flight}. Modification may change flight, cabin, added-bag quantity, or payment; \code{add\_baggage} defaults 0 and adds bags; omitted new\_payment\_method reuses original.

\TBMReportHeading{Task behavior}
To change an existing return flight without reservation ID, obtain/ask user ID and look up reservations. If multiple bookings appear, choose an unambiguous matching origin/destination return leg, otherwise ask confirmation. If no new flight is given, search the route and choose an available evening departure on the same calendar date as the existing return flight, excluding later dates.
\par\vspace{4pt}
After selecting the flight, call \code{modify\_flight} with user\_id, reservation\_id, selected flight number, and only requested optional changes. Preserve payment by omitting new\_payment\_method or passing existing only if explicitly requested; preserve cabin by omitting/passing existing; preserve checked bags by omitting add\_baggage or 0. If asked to confirm, retrieve and report updated route, times, cabin, baggage, and payment from tool output.

\TBMReportHeading{Failure behavior}
Invented or unlooked-up new\_flight\_no calls were rejected as invalid or route-mismatched; recovery came only after searching and retrying with a returned same-day evening flight. The tool cannot edit departure time directly; select another flight number. Password collection was unnecessary.
\par\vspace{4pt}
In the same-evening return change, passing a non-original new\_cabin changed cabin/seat type and triggered a refund, which the user rejected. Later retry omitting cabin/payment and reporting no fee did not prove cabin restoration or refund/payment reversal without explicit restoration request or lookup.

\end{tcolorbox}
\caption{An example domain entry from the final TBM report.}
\label{fig:tbm_report_case}
\end{figure}
\endgroup

%% file: figures/fixed_env_training_rewards_float.tex
\begin{figure}[htbp]
\begin{minipage}{\linewidth}
\centering
\includegraphics[width=\linewidth]{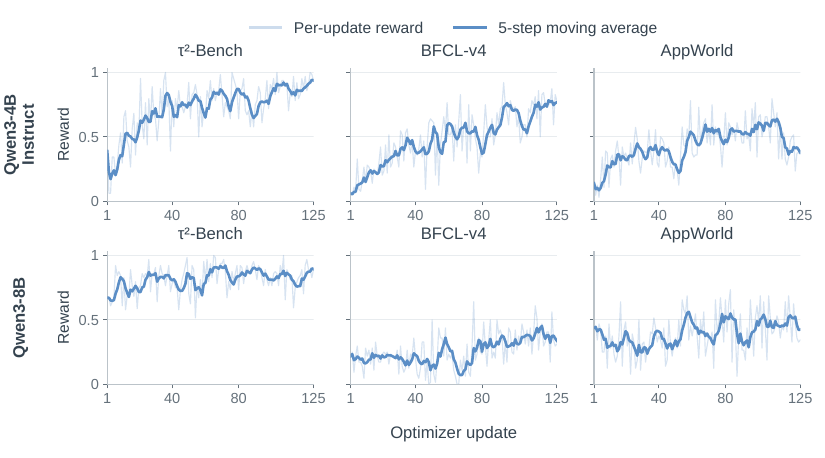}
\caption{Training rewards for the 8 \texttt{Fixed-Env} experiments.
Light lines show the mean reward per optimizer update; dark lines show
a 5-update moving average.}
\label{fig:fixed_env_training_rewards}
\end{minipage}
\end{figure}